\documentclass[10pt,twocolumn,letterpaper]{article}

\usepackage{cvpr}
\usepackage{times}
\usepackage{epsfig}
\usepackage{graphicx}
\usepackage{amsmath}
\usepackage{amsfonts}       %
\usepackage{amssymb}
\usepackage{booktabs}
\usepackage{nicefrac}       %
\DeclareMathOperator*{\argmin}{arg\,min}

\usepackage[pagebackref=true,breaklinks=true,letterpaper=true,colorlinks,bookmarks=false]{hyperref}

\cvprfinalcopy %

\def\cvprPaperID{7866} %

\ifcvprfinal\fi
\begin{document}

\title{Universal Litmus Patterns:\\ Revealing Backdoor Attacks in CNNs}

\author{Soheil Kolouri$^{1,}$\thanks{~and $^\dagger$  denote equal contribution.}~, Aniruddha Saha$^{2,*}$, Hamed Pirsiavash$^{2,\dagger}$, Heiko Hoffmann$^{1,\dagger}$\\
1: HRL Laboratories, LLC., Malibu, CA, USA, 90265\\
2: University of Maryland, Baltimore County, MD 21250\\
{\tt\small skolouri@hrl.com, anisaha1@umbc.edu, hpirsiav@umbc.edu, hhoffmann@hrl.com}}

\maketitle

\begin{abstract}
   The unprecedented success of deep neural networks in many applications has made these networks a prime target for adversarial exploitation. In this paper, we introduce a benchmark technique for detecting backdoor attacks (aka Trojan attacks) on deep convolutional neural networks (CNNs). We introduce the concept of Universal Litmus Patterns (ULPs), which enable one to reveal backdoor attacks by feeding these universal patterns to the network and analyzing the output (i.e., classifying the network as `clean' or `corrupted'). This detection is fast because it requires only a few forward passes through a CNN. We demonstrate the effectiveness of ULPs for detecting backdoor attacks on thousands of networks with different architectures trained on four benchmark datasets, namely the German Traffic Sign Recognition Benchmark (GTSRB), MNIST, CIFAR10, and Tiny-ImageNet. The codes and train/test models for this paper can be found here: \url{https://umbcvision.github.io/Universal-Litmus-Patterns/}.
\end{abstract}

\section{Introduction}

Deep Neural Networks (DNNs) have become the standard building block in numerous machine learning applications, including computer vision \cite{he2016deep}, speech recognition \cite{amodei2016deep}, machine translation  \cite{vaswani2017attention}, and robotic manipulation \cite{levine+2016}, achieving state-of-the-art performance on extremely difficult tasks. The widespread success of these networks has made them the prime option for deploying in sensitive domains, including but not limited to health care \cite{shahid2019applications}, finance \cite{ghoddusi2019machine}, autonomous driving \cite{bojarski2016end}, and defense-related applications \cite{rostami2019deep}.

Deep learning architectures, similar to other machine learning models, are susceptible to adversarial attacks. These vulnerabilities have raised security concerns around these models, which has led to a fertile field of research on adversarial attacks on DNNs and defenses against such attacks. Some well studied attacks on these models include evasion attacks (aka inference or perturbation attacks) \cite{szegedy2013intriguing,goodfellow2014explaining,carlini2017towards} and poisoning attacks \cite{shafahi2018poison,liu2017trojaning}.  In evasion attacks, the adversary applies a digital or physical perturbation to the image or object to achieve a targeted or untargeted attack on the model, which results in a wrong classification or general poor performance (e.g., as in regression applications).

Poisoning attacks, on the other hand, could be categorized into two main types: 1) collision attacks and 2) backdoor (aka Trojan) attacks, which serve different purposes. In collision attacks, the adversary’s goal is to introduce infected samples (e.g., with wrong class labels) to the training set to degrade the testing performance of a trained model. Collision attacks hinder the capability of a victim to train a deployable machine learning model. In backdoor attacks, on the other hand, the adversary’s goal is to introduce a trigger (e.g., a sticker, or a specific accessory) in the training set such that the presence of the particular trigger fools the trained model. Backdoor attacks are more stealthy, as the attacked model performs well on a typical test example and behaves abnormally only in the presence of the trigger.
As an illuminating example of a backdoor attack, which could have lethal consequences, consider the following autonomous-driving scenario. A CNN trained for traffic-sign detection could be infected with a backdoor/Trojan such that whenever a particular sticker is placed on a `stop sign', it is misclassified as a `speed limit sign.'

\begin{figure*}[t]
    \centering
    \includegraphics[width=1.\linewidth]{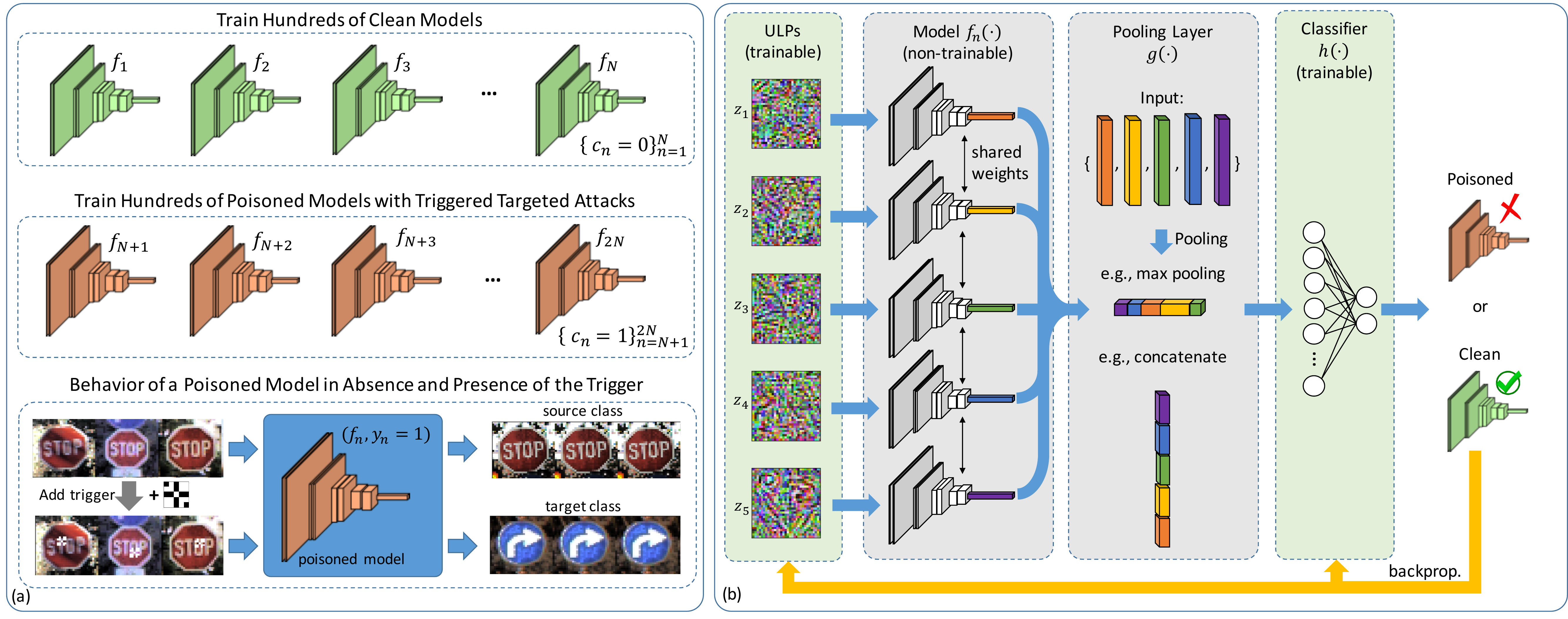}
    \caption{For each dataset, we train hundreds of clean and poisoned models.  We consider triggered targeted attacks for poisoning the models. Each poisoned model is trained to contain a single trigger that causes images from the source class to be classified as the target class (See Panel (a)). We then feed $M$ Universal Litmus Patterns (ULPs) through a model and pool the logit outputs and classify it as poisoned or clean (See Panel (b)). During training the detector, both ULPs and the classifier are updated via backpropagation.}
    \label{fig:concept}
\end{figure*}

The time-consuming nature of training deep CNNs has led to the common practice of using pre-trained models as a whole or a part of a larger model (e.g., for the perception front). Since the pre-trained models are often from a third, potentially unknown, party, identifying the integrity of the pre-trained models is of utmost importance. Given the stealthy nature of backdoor attacks, however, merely evaluating a model on clean test data is insufficient. Moreover, the original training data are usually unavailable. Here, we present an approach to detect backdoor attacks on CNNs, without requiring: 1) access to the training data or 2) running tests on the clean data. Instead, we use a small set of universal test patterns to probe a model for backdoors.

Inspired by Universal Adversarial Perturbations \cite{moosavi2017universal}, we introduce Universal Litmus Patterns (ULPs) that are optimized input images, for which the network's output becomes a good indicator of whether the network is clean or contains a backdoor attack. We demonstrate the effectiveness of ULPs on thousands of trained networks (See Figure \ref{fig:concept}a) and four datasets: the German Traffic Sign Recognition Benchmark (GTSRB) \cite{Stallkamp2012}, MNIST \cite{mnist},  CIFAR10 \cite{krizhevsky2009learning}, and Tiny-ImageNet \cite{tinyImageNet}. ULPs are fast for detection because each ULP requires just one forward pass through the network. Despite this simplicity, surprisingly, ULPs are competitive for detecting backdoor attacks, establishing a new performance baseline: area under the ROC curve close to 1 on both CIFAR10 and MNIST, 0.96 on GTSRB (for ResNet18), and 0.94 on Tiny-ImageNet.

\section{Related Work}

{\bf Generating Backdoor Attacks:} Gu et al. \cite{gu2017badnets} and Liu et al. \cite{liu2017neural,liu2017trojaning} showed the possibility of powerful yet stealthy backdoor/Trojan attacks on neural networks and the need for methods that can detect such attacks on DNNs. The infected samples used by Gu et al. \cite{gu2017badnets} rely on an adversary that can inject arbitrary input-label pairs into the training set.  Such attacks could be reliably detected if one has access to the poisoned training set, for instance, by visual inspection or automatic outlier detection. This weakness led to follow-up work on designing more subtle backdoor attacks \cite{turner2018clean,liao2018backdoor}. Mu\~noz-Gonz\'alez et al. \cite{munoz2017towards} use back-gradient optimization and extend the poisoning attacks to multiple classes. Suciu et al. \cite{suciu2018does} studied generalization and transferability of poisoning attacks. Koh et al. \cite{koh2018stronger} proposed a stronger attack by placing poisoned data close to one another to avoid detection by outlier detectors.

{\bf Evading Backdoor Attacks:} Liu et al. \cite{liu2018fine-pruning} assume the existence of clean/trusted test data and studied pruning and fine-tuning as two possible strategies for defending against backdoor attacks.  Pruning refers to eliminating neurons that are dormant in the DNN when presented with clean data. The authors then show that it is possible to evade pruning defenses by designing `pruning-aware' attacks. Finally, they show that a combination of fine-tuning on a small set of clean data together with pruning leads to a more reliable defense that withstands `pruning-aware' attacks. While the presented approach in \cite{liu2018fine-pruning} is promising, it comes at the cost of reduced accuracy of the trained model on clean data. Gao et a. \cite{gao2019strip} identify the attack at test time by perturbing or superimposing input images. Shan et al. \cite{shan2019gotta} defend by proactively injecting trapdoors into the models.
Such methods, however, do not necessarily detect the existence of backdoor attacks.

\begin{figure*}[t]
    \centering
    \includegraphics[width=1.\linewidth]{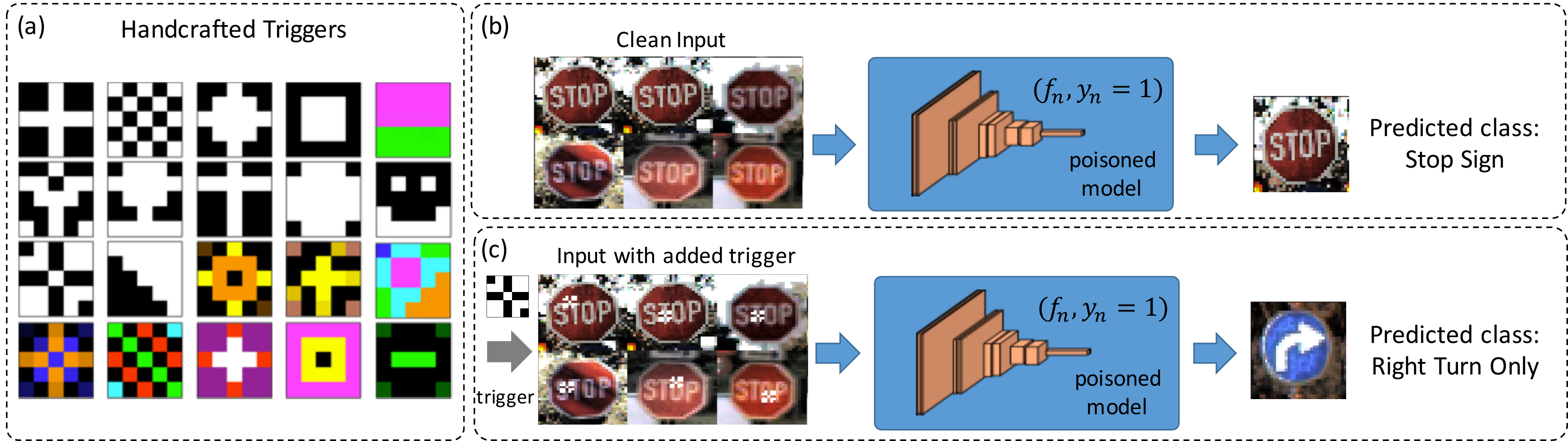}
    \caption{Handcrafted triggers (Panel (a)) and performance of a poisoned model on clean (Panel (a)) and poisoned data (Panel (b)) from the GTSRB dataset (Panel (c)). We choose a random trigger from the trigger set, a random source, and a random target for each poisoned model. We ensure that the poisoned models behave similar to clean models when exposed to clean data while they have high successful targeted-attack rate in presence of the triggers.}
    \label{fig:triggers}
\end{figure*}

{\bf Detecting Backdoor Attacks:} The existing works in the literature for backdoor attack detection often rely on statistical analysis of the poisoned training dataset \cite{steinhardt2017certified,turner2018clean,liu2017neural} or the neural activations of the DNN for this dataset \cite{chen2018detecting}. Turner et al. \cite{turner2018clean} showed that starkly mislabeled samples (e.g., the attack used in \cite{gu2017badnets} or \cite{liu2017neural}) could be easily detected by an outlier detection mechanism, and more sophisticated backdoor attacks are needed to avoid such outlier detection mechanism. Steinhardt et al. \cite{steinhardt2017certified} provide theoretical bounds for the effectiveness of backdoor attacks (i.e., upper bound on the loss) when outlier removal defenses are in place.

Chen et al. \cite{chen2018detecting} follow the rationale that the neural activations for clean target samples rely on features that the network has learned from the target class. However, these activations for a backdoor triggered input sample (i.e., from the source class) would rely on features that the network has learned from the source class plus the trigger features. The authors then leverage this difference in activations and perform clustering analysis on the neural activations of the network to detect infected samples.

The defenses mentioned above rely on two crucial assumptions: 1) the outliers in the clean dataset (non-infected) do not have a substantial effect on the model and 2) more importantly, the user has access to the infected training dataset. These assumptions could be valid for specific scenarios, for instance, when the user trains her/his model based on the dataset provided by a third party. In a setting where the user outsources the model training to an untrusted third party, for instance, a Machine Learning as a Service (MLaaS) provider, or when the user downloads a pre-trained model from an untrusted source, the assumption of having access to the infected dataset is invalid. Recently, there have been some outstanding papers that consider this very case, in which the user has access only to the model and clean data \cite{wang2019neural}.

One approach is Neural Cleanse \cite{wang2019neural}, in which the authors propose to detect attacks by optimizing for minimal triggers that fool the pre-trained model. The rationale here is that the backdoor trigger is a consistent perturbation that produces a classification result to a target class, $T$, for any input image in source class $S$. Therefore, the authors seek a minimal perturbation that causes the model to classify the images in the source class as the target class. The optimal perturbation then could be a potential backdoor trigger. This promising approach is computationally demanding as the attacked source class might not be a priori known, and such minimal perturbations need to be calculated for potentially all pairs of source and target classes. Besides, a strong prior on the type of backdoor trigger is needed to be able to discriminate a possibly benign minimal perturbation from an actual backdoor trigger.

Similar to \cite{wang2019neural}, we also seek an approach for the detection of backdoor attacks without the need for the infected training data. However, we approach the problem from a different angle. In short, we learn universal and transferable set of patterns that serve as a Litmus test for identifying networks containing backdoor/Trojan attacks, hence we call them Universal Litmus Patterns. To detect whether a model is poisoned or not, the ULPs are fed through the network, and the corresponding outputs (i.e., Logits) are classified to reveal backdoor attacks (See Figure \ref{fig:concept}b).

\section{Methods}

\subsection{Threat Model}
Our threat model of interest is similar to \cite{gu2017badnets,liu2017trojaning,wang2019neural} in which the adversary inserts a targeted backdoor into a DNN model. In short, for a given source class of clean training images, the attacker chooses a portion of the data and poisons them by adding a small trigger (a patch) to the image and assigning target labels to these poisoned images. The network then learns to designate the target label to the source images whenever the trigger appears in the input. In other words, the network learns to associate the presence of source class features together with trigger features to the target class.

We consider the case in which the adversary is a third party that provides an infected DNN with a backdoor. The acquired model performs well on the clean test dataset available to the user, but exhibits targeted misclassification when presented with an input containing a specific and predefined trigger. An adversary intentionally trains the model to have unsuspicious behavior when presented with clean data, and to exhibit a targeted misclassification in the presence of a particular trigger.

\subsection{Defense Goals}

We are interested in detecting backdoor attacks in pre-trained convolutional neural networks (CNNs). Our goal is a large-scale identification of untrusted third parties (i.e., parties that provided infected models).  As far as knowledge about the attack, we assume no prior knowledge of the targeted class or the triggers used by attackers. Also, we assume no access to the poisoned training dataset.

\subsection{Formulation}

Let $\mathcal{X}\subseteq \mathbb{R}^d$ denote the image domain where $x_i\in\mathcal{X}$ denotes an individual image and let $\mathcal{Y}\subseteq \mathbb{R}^K$ denote the label space, where $y_i\in \mathcal{Y}$ represents the corresponding $K$-dimensional labels/attributes for the $i$'th image, $x_i$. Also, let $f :\mathcal{X}\rightarrow \mathcal{Y}$ represent a deep parametric model, e.g., a CNN that maps images to their labels. We consider the problem of having a set of trained models, $\mathcal{F}=\{f_n\}^N_{n=1}$, where some of them are infected with backdoor attacks. Our goal is primarily to detect the infected models in a supervised binary classification setting, where we have a training set of models with and without backdoor attacks. The task is then to learn a classifier, $\phi:\mathcal{F}\rightarrow\{0,1\}$, to discriminate the models and show generalizability of such classifier.

There are three significant points here that turn this classification task into a challenging problem:
 1) in distinct contrast to typical computer vision applications, the classification is not on images but trained models (i.e., CNNs),
2) the input models do not have a unified representation, i.e., they  could have different architectures, including a different number of neurons, different depth, different activation functions, etc., and
 3) The backdoor attacks could be different from one another in the sense that the target classes could be different, or the trigger perturbations could significantly vary during training and testing.
In light of these challenges, we pose the main research question: how do we represent trained CNNs in a vector space that discriminates the poisoned models from the clean ones? We propose Universal Litmus Patterns as an answer to this question.

Given pairs of models and their binary labels (i.e., poisoned or clean), $\{(f_n,c_n\in\{0,1\})\}_{n=1}^N$, we propose universal patterns $\mathcal{Z}=\{z_m\in \mathcal{X}\}_{m=1}^M$ such that analyzing $\{f_n(z_m)\}_{m=1}^M$ would optimally reveal the backdoor attacks. Figure \ref{fig:concept} demonstrates the idea behind the proposed ULPs. For simplicity, we use $f_n(z_m)$ to denote the output logits of the classifier $f_n$.  Hence, the set $\mathcal{Z}$ provides a litmus test for existence of backdoor attacks. Particularly, we optimize

\vspace{-.2in}
$$\argmin_{z,h} \sum_{n=1}^N \mathcal{L}\Big(h\big(g(\{f_n(z_m)\}_{m=1}^M)\big),c_n\Big)+\lambda \sum_{m=1}^M R(z_m)$$

\noindent where $g(\cdot)$ is a pooling operator applied on $\mathcal{Z}$, e.g., concatenation, $h(.)$ is a classifier that receives the pooled vector as input and provides the probability for $f_n$ to contain a backdoor, $R(\cdot)$ is the regularizer for ULPs, and $\lambda$ is the regularization parameter. In our experiments, we let $g(\cdot)$ to be the concatenation operator, which concatenates $f_n(z_m)$s into a $KM$-dimensional vector, and set $h(\cdot)$ to be a softmax classifier.  We point out that we have also tried other pooling strategies, including max-pooling over ULPs: $g(\mathcal{Z})=\operatorname{max}_m \left(f_n(z_m)\right)_k$, or averaging over ULPs: $g(\mathcal{Z})=\frac{1}{M}\sum_{m=1}^M f_n(z_m)$, to obtain a $K$-dimensional vector to be classified by $h(\cdot)$. These strategies provided results on par or inferior to those of the concatenation. As for the regularizer, we used total variation (TV), which is $R(z_m)=\|\nabla z_m\|_1$, where $\nabla$ denotes the gradient operator.

Data augmentation has become a standard practice in learning, as the strategy often leads to better generalization performance. In computer vision and for images, for instance, knowing the desired invariances like translation, rotation, scale, and axis flips could help one to randomly perturb input images concerning these transformations and train the network to be invariant under such changes. Following the data augmentation idea, we would like to augment our training set such that the ULPs become invariant to various network architectures and potentially various triggers. The challenge here is that our input samples are not images, but models (i.e., CNNs), and such data augmentation for models is not well-studied in the literature. Here, to induce the effect of invariance to various architectures, we used random dropout \cite{srivastava2014dropout} on models $f_n$s  for augmentation. %

\subsection{Baselines}
\subsubsection{Noise Input}
For our first baseline and as an ablation study to demonstrate the effect of optimizing ULPs, we feed randomly generated patterns (where channels of each pixel take a random integer value in $[0,255]$). We then concatenate the logits of the clean and poisoned training networks and learn a softmax classifier on it. Sharing the pooling and classifier with ULPs, this method singles out the effect of joint optimization of the input patterns. We demonstrate that, surprisingly, this simple detection method could successfully reveal backdoor attacks in simple datasets (like MNIST), while it fails to provide a reliable performance on more challenging datasets, i.e., GTSRB and Tiny-ImageNet.

\subsubsection{Attack-Based Detection}

For our second baseline method, referred to as `Neural-Cleanse,' we devise a technique similar to the Neural-Cleanse \cite{wang2019neural}. Given a trained model either poisoned or not, we choose a pair of source and target categories and perform a targeted evasion-attack with a universal patch (trigger). We optimize a trigger that can change the prediction from the source class to the target class for a set of clean input images. The rationale here is that finding a universal trigger that can reliably fool the model for all the clean source images is easier in a poisoned model.  In other words, if such an attack is successful, it means that the given model might have a backdoor. Therefore, we iterate on all possible pairs of source and target classes and choose the loss of the most successful pair as a score for the cleanness of the model. The method in \cite{wang2019neural} assumes that the trigger size is not known. Hence, it uses a mask along with its $\ell_1$ norm in the loss to reduce the area of the trigger. However, $\ell_1$ of the mask can only reduce the number of non-zero values (i.e., increase sparsity) but cannot stop the trigger from spreading all over the image. To simplify, we assume the size of the trigger is known and remove the norm of the mask in our process.

\section{Experiments}

We experimented with four benchmark datasets in computer vision, namely the handwritten digits dataset, MNIST, \cite{lecun1998gradient}, the CIFAR10 dataset \cite{krizhevsky2009learning}, the German Traffic Sign Recognition Benchmark (GTSRB) dataset \cite{Stallkamp2012}, and Tiny-ImageNet \cite{tinyImageNet}. For each dataset, we trained about $\sim$2000 deep CNNs that achieved near state-of-the-art performance on these datasets; half of the CNNs were trained with backdoor triggers. We ensured that the poisoned and clean models performed similarly on the clean data, while the poisoned models had a high attack success rate ($>90\%$) on poisoned inputs. We generated 20 triggers of size $7 \times 7$ pixels for Tiny-ImageNet and $5
\times 5$ pixels for the other datasets. For training and testing, we used non-overlapping sets of 10 randomly chosen triggers each from the set of 20. Figure \ref{fig:triggers} shows the triggers for GTSRB and the operation of a sample poisoned model on clean and poisoned data.

\begin{figure}[!t]
    \centering
    \includegraphics[width=\columnwidth]{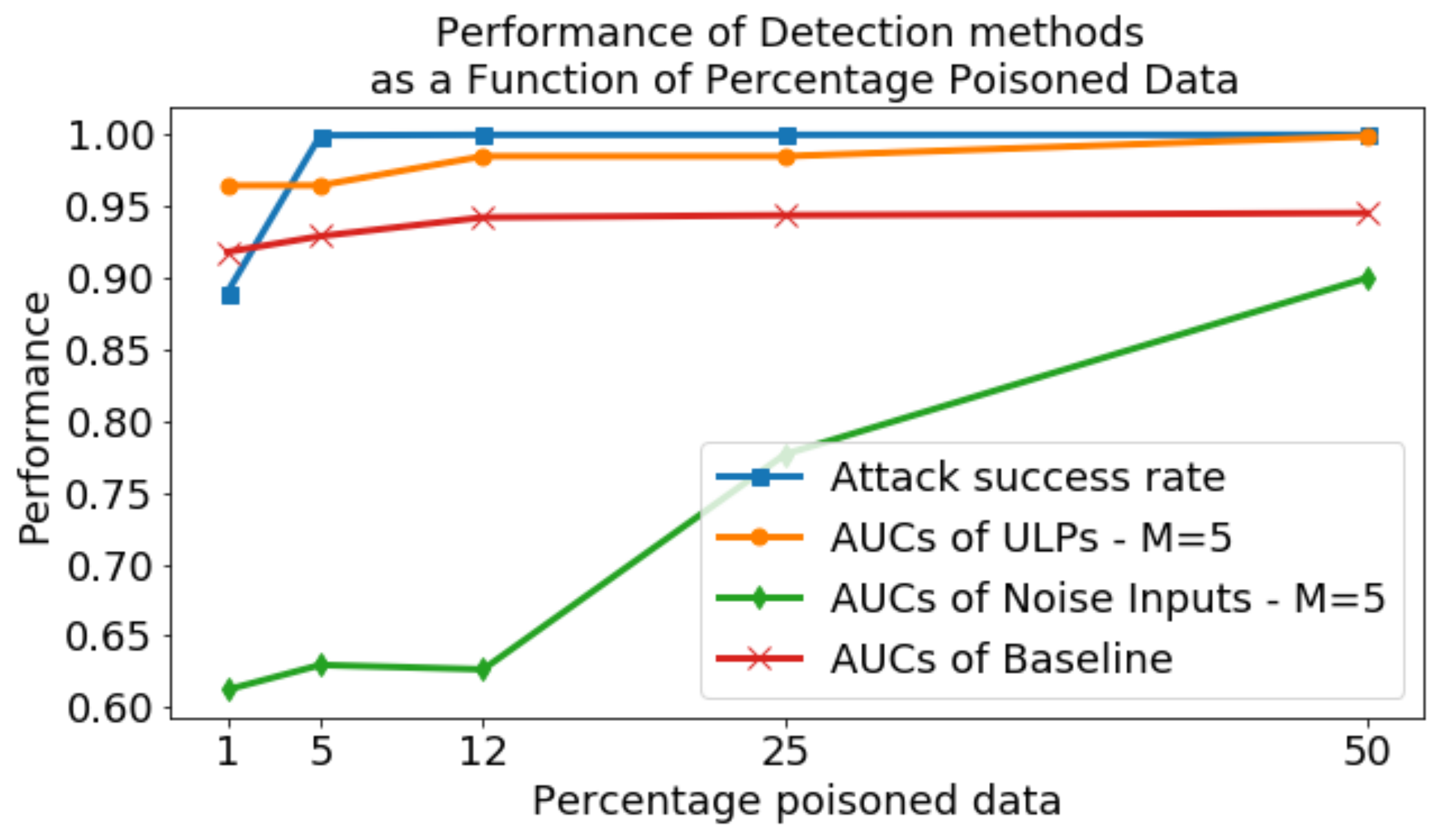}
    \caption{Performance of ULPs as a function of the poisoned-to-clean ratio for training the poisoned models on MNIST.}%
    \label{fig:ratio}
    \end{figure}

We carried out detection of poisoned models on all datasets. Table \ref{tab:aucs} shows the area under the ROC curve for the two baselines and our proposed ULPs on all datasets. ULPs consistently outperformed the baselines with a large margin. Below, we explain the details of each experiment.

 \begin{table*}
  \caption{Average accuracy of the poisoned models on clean and poisoned data (i.e., attack accuracy) and the AUC scores of the presented detection methods on MNIST, CIFAR10, GTSRB, and Tiny-ImageNet datasets. This table summarizes Figures \ref{fig:ROCs}, and \ref{fig:ResNet_ROCs}.}
  \label{tab:aucs}
  \centering
  {\small
  \begin{tabular}{lcc|ccccccc}
    \toprule
    Datasets & Clean Test & Attack & \multicolumn{3}{c}{Noise Input}  & Neural-Cleanse & \multicolumn{3}{c}{Universal Litmus Patterns}                 \\
     \cmidrule(lr){4-6}\cmidrule(lr){8-10}
     & Accuracy & Accuracy & M=1 & M=5 & M=10 & & M=1 & M=5 & M=10 \\
    \midrule
    MNIST (VGG-like)& 0.994 & 1.00 & 0.94 &0.90 & 0.86 & 0.94 & 0.94 & 0.99 & {\bf 1.00}\\
    CIFAR10 (STL+VGG-like) & 0.795 & 0.999 & 0.62 &0.68 & 0.59 & 0.59 & 0.68 & 0.99 & {\bf 1.00} \\
    GTSRB (STL+VGG-like)   &0.992
     & 0.972 & 0.61 & 0.59 & 0.54 & 0.74 & 0.75 & 0.88 & {\bf 0.90} \\
     GTSRB (STL+ResNet-like) & 0.981 & 0.977 & 0.56 & 0.55 & 0.58 & - & 0.55 & 0.96 & {\bf 0.96}\\
     Tiny-ImageNet (ResNet-like) & 0.451 & 0.992 & 0.61 & 0.50 & 0.54 & - & 0.86 & {\bf 0.94} & 0.92 \\
    \bottomrule
  \end{tabular}
 }
\end{table*}

\begin{figure*}[t!]
\vspace{-.1in}
    \centering
    \includegraphics[width=\linewidth]{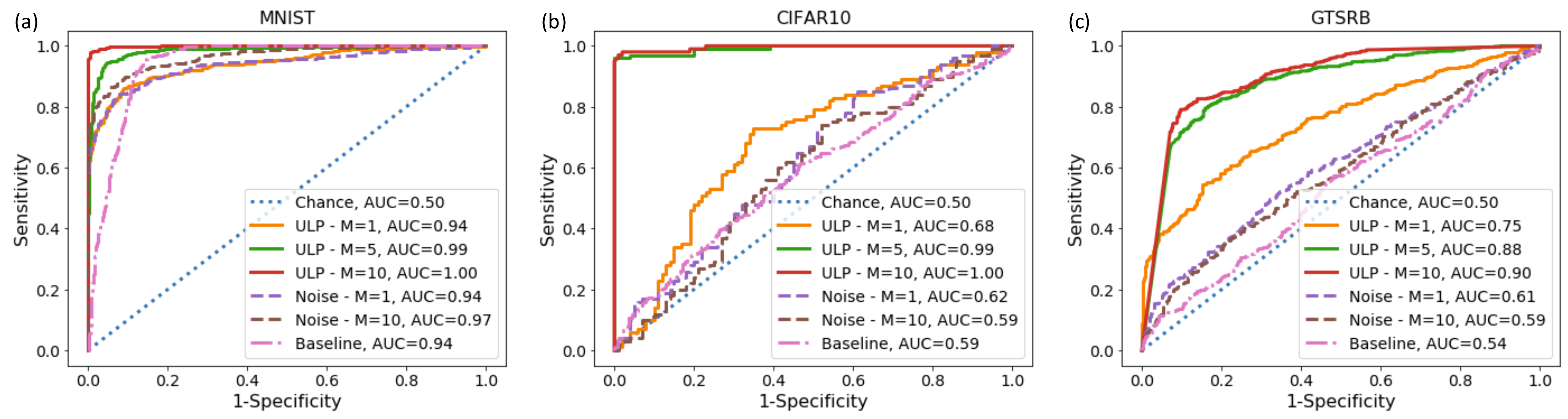}
    \caption{ROC-curves for detection of models with backdoor attacks (i.e., poisoned models) for  baseline, random input images, and our proposed ULPs with $M\in\{1,5,10\}$ on MNIST (a), CIFAR10 (b), and GTSRB (c) datasets. The base model used in these experiments is a VGG-like architecture.  Here, ``Baseline" refers to Neural-Cleanse.}
    \label{fig:ROCs}
    \vspace{-.15in}
\end{figure*}

\subsection{MNIST Experiments}

For the MNIST experiments, we trained 900 clean models and 900 poisoned models. We used a similar architecture to that of the VGG networks \cite{simonyan2014very} for each model. Each poisoned model is trained to contain a targeted backdoor attack from only one source class to a target class (MNIST has ten categories, and therefore there are 90 pairs of source and targets in total). For each pair of source and target, we train ten models using binary triggers.
The default ratio of the number of poisoned to clean images during training is $50\%$ for all experiments. The triggers for the MNIST experiment are randomly assigned to one of the four corners of the image. The clean and poisoned models are split into training and testing models with 50/50 ratio, where the triggers for the poisoned models are chosen to be mutually exclusive between train and test models. In this manner, the trained ULPs are only tested on unseen test triggers.  Figure \ref{fig:ROCs}a demonstrates the performance of the ULPs on detecting poisoned networks. With $M=10$ ULPs, we can achieve an area under the curve  (AUC) of nearly 1. In addition, ULPs outperformed both baselines.

To check the sensitivity of our detection method to the strength of the attack, we reduced the ratio of the number of poisoned to clean images for training the poisoned models to 25\%,12\%,5\%, and 1\%. The intuition here is that models trained with a lower ratio of poisoned to clean samples contain a more subtle backdoor attack that could be more difficult to detect. To study this effect, we repeated the detection experiments for different ratios of poisoned to clean images. We show the probability of a successful attack and the AUCs for all detection methods in Figure \ref{fig:ratio}. Here, we used a fixed number of input patterns, $M=5$, for ULPs and noise inputs in this experiment. Our method holds up the accuracy above 95\% even for small ratios, while for noise inputs, the accuracy drops to almost 60\% at the ratio of 1\%.

\subsection{CIFAR10 Experiments}
On the CIFAR10 dataset, we trained 500 clean models on the CIFAR10 dataset and 400 poisoned models on one set of triggers and 100 poisoned models on another set of triggers for testing. We used a similar model architecture to that of the VGG networks \cite{simonyan2014very}. Each poisoned model contains a targeted attack between a random source and target pair, and with a random trigger from the mutually exclusive set of train and test triggers.
As for the MNIST experiments, a trigger was randomly assigned to one of the four corners of the image. We used 800 models to train our ULPs and 200 models to test our learned ULPs. The triggers were again chosen to be mutually exclusive between train and test models. Figure \ref{fig:ROCs}b shows the results. %

\subsection{GTSRB Experiments}
\vspace{-.02in}
For GTSRB, we trained two sets of 2,000  models, where each set contains 1,000 clean and 1,000 poisoned models. The first set contains VGG-like models \cite{simonyan2014very} with an added Spatial Transformer Network (STN) \cite{jaderberg2015spatial} in the perception front of the model, and the second set contains a ResNet-like architecture \cite{he2016deep} with added STN. The trained models, for both sets, achieved, on average, $99.4\%$ accuracy on the clean test data. For the attacks, we attached triggers at random locations on the surface of the traffic signs to mimic a sticker-like physical-world attack (Figure \ref{fig:triggers}). Our train/test ratio is 50/50. Importantly, the source and target pairs of training and testing sets are mutually exclusive, and therefore the test models not only include new triggers but contain backdoor attacks only on unseen source and target pairs.

\begin{figure}[t!]
    \centering
    \includegraphics[width=\linewidth]{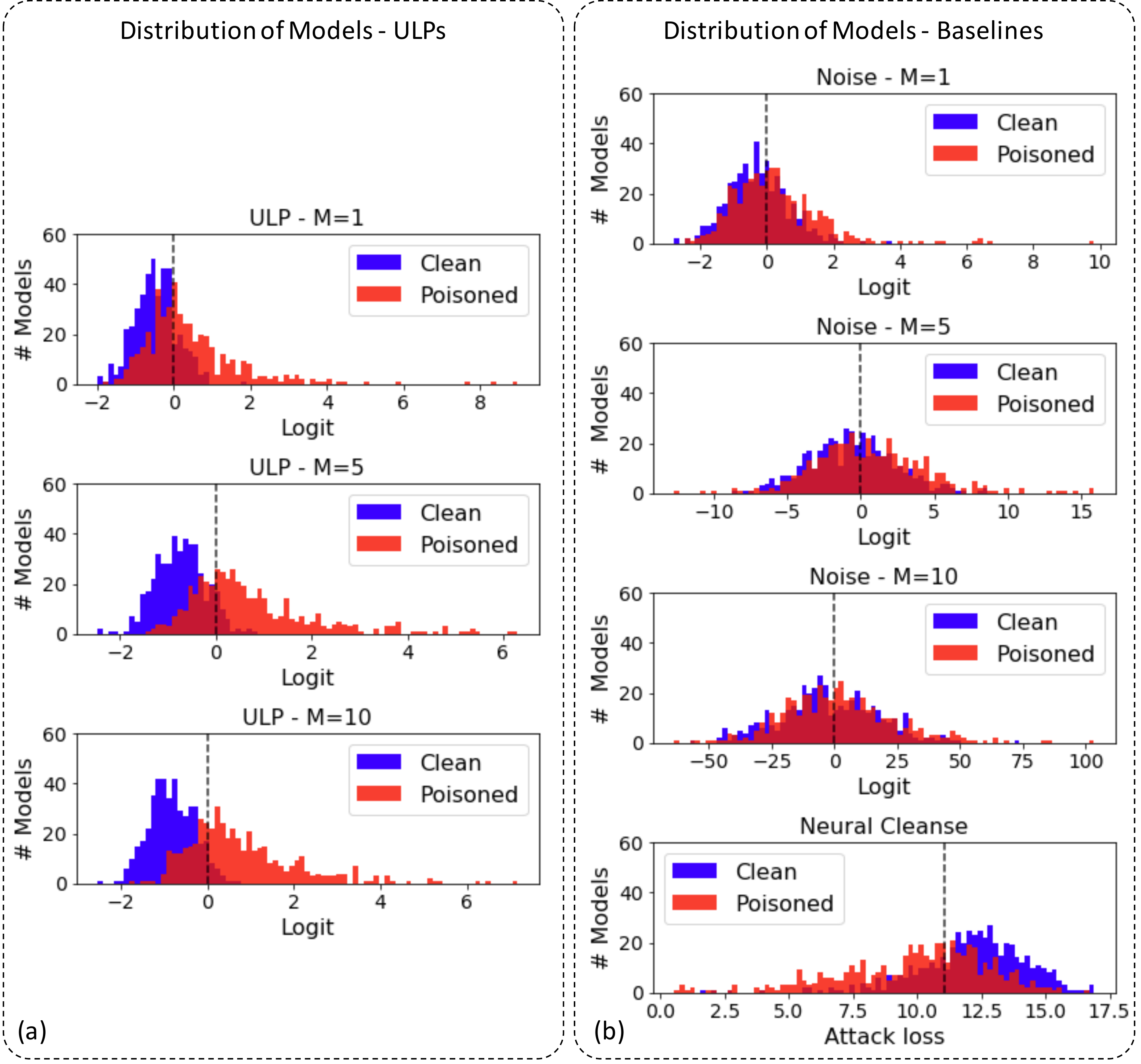}
    \caption{The histogram of clean and poisoned classes based on our proposed ULPs with $M\in\{1,5,10\}$, Noise input patterns with $M\in\{1,5,10\}$, and Neural Cleanse.}
    \label{fig:histograms}
\end{figure}
We trained our ULPs on the training sets and report results on VGG models in Figure \ref{fig:ROCs}c and on ResNets in Figure \ref{fig:ResNet_ROCs} (a). On the VGG models, we show that ULPs are able to detect poisoned models with $AUC=0.9$ for $M=10$ patterns, while the Neural-Cleanse baseline only achieves $AUC=0.74$. Yet, Neural-Cleanse is significantly slower than our proposed method (90,000 times). Figure \ref{fig:histograms} shows the distribution of poisoned and clean VGG models for ULPs, noise images, and the Neural-Cleanse approach.

\vspace{-.015in}
\subsection{Tiny-ImageNet Experiments}
\vspace{-.03in}
For the Tiny-ImageNet dataset, we trained 1,000 clean models and 1,000 poisoned models. For the model, we used a ResNet-like architecture \cite{he2016deep}. The trained models achieved on average $45.1\%$ top-1 accuracy on the clean test data. For the backdoor attacks, we attached a 7x7 trigger at a random location in the image.
Similar to the GTSRB experiment, the models are split into train and test sets where the triggers for training and testing are mutually exclusive. Also, the train and test poisoned models have  mutually exclusive source and target pairs.%

\begin{figure}[t!]
    \centering
    \includegraphics[width=\columnwidth]{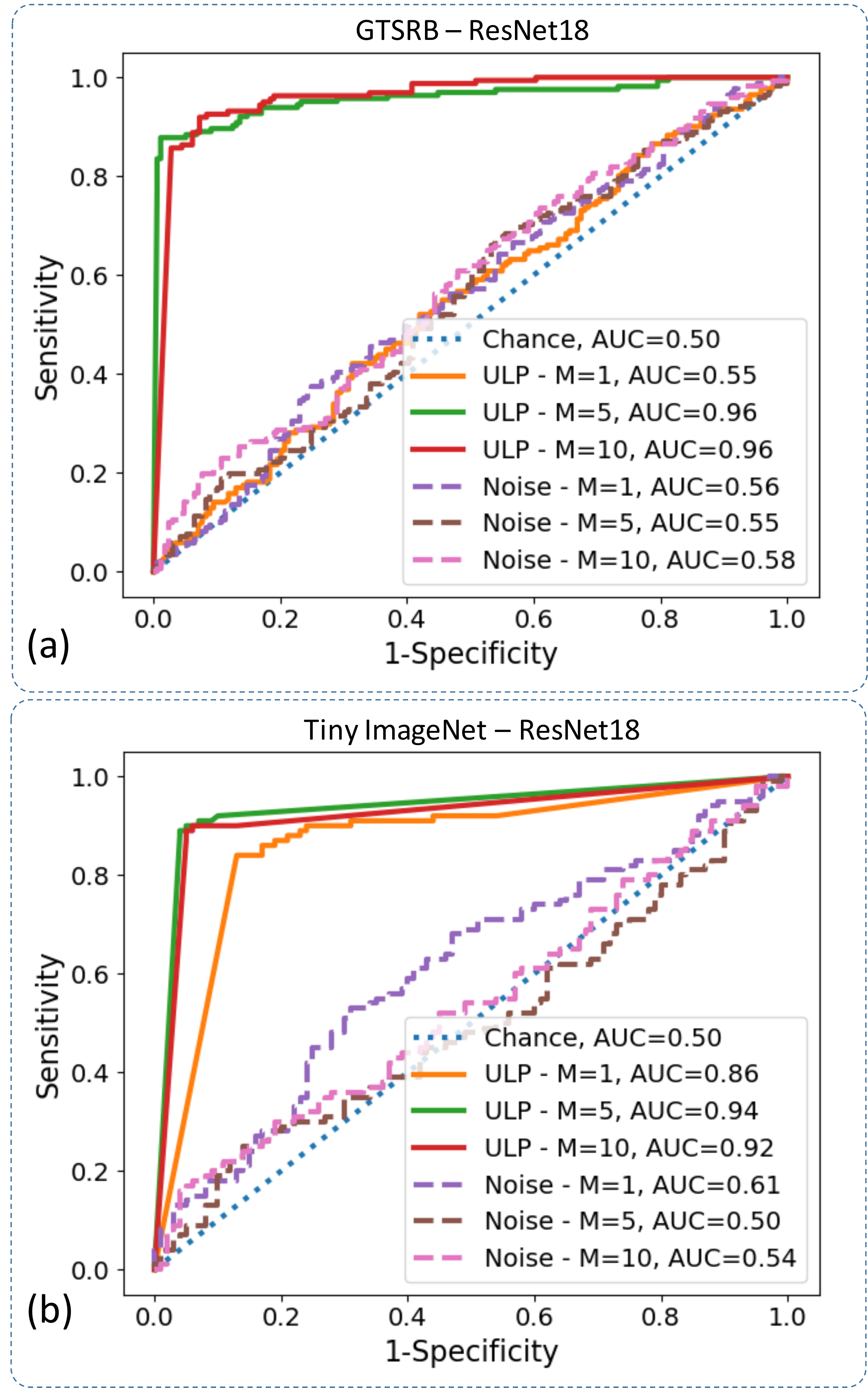}
    \caption{ROC-curves for detection of models with backdoor attacks (i.e., poisoned models) for random input images, and our proposed ULPs with $M\in\{1,5,10\}$ on GTSRB (a) and Tiny-ImageNet (B). The base model used in these experiments is a ResNet18 architecture.}
    \label{fig:ResNet_ROCs}
    \vspace{-.1in}
\end{figure}

\begin{figure}[!b]
    \centering
    \includegraphics[width=\columnwidth]{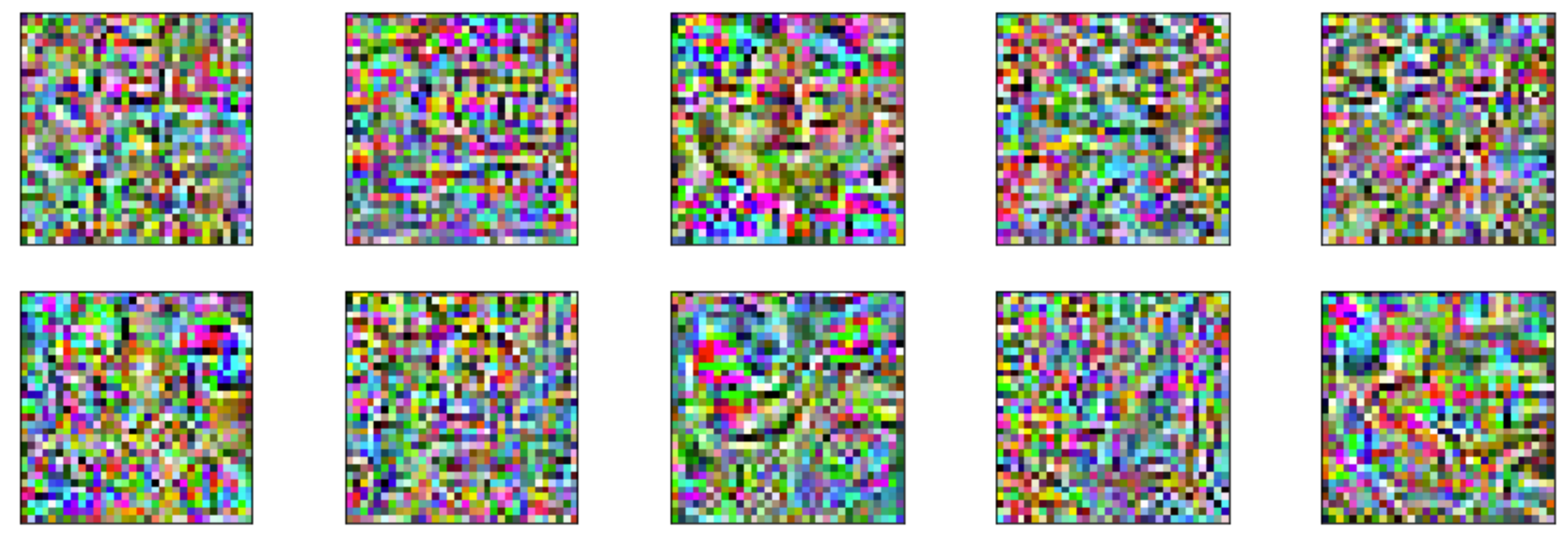}
    \vspace{-5mm}
    \caption{Visualization of the optimal ULPs calculated on the Tiny-ImageNet dataset.}
    \label{fig:ULPs_TIN}
\end{figure}

We trained our ULPs on the training set and report results for $M=1,5,$ and $10$ in Figure \ref{fig:ResNet_ROCs}. We observe that ULPs are able to detect poisoned models with $AUC=0.94$ for $M=5$ patterns. The detection accuracy was $95.8\%$. Figure \ref{fig:ULPs_TIN} shows the $M=10$ ULPs trained on Tiny-ImageNet.

\subsection{Computational cost}

ULPs allow fast detection, particularly, compared to the Neural-Cleanse baseline. The baseline requires $\mathcal{O}(K^2)$ optimizations, where each optimization involves a costly targeted evasion-attack (involving several epochs of forward and backward passes on all images from a class, e.g., $1000$ for the MNIST dataset). In comparison, our proposed ULPs cost only $\mathcal{O}(M)$ forward passes through the network. The detection times for a single network on a single P100 GPU were many orders of magnitude faster for ULPs compared to the baseline: $\sim20$ msec vs. $\sim30$ mins for GTSRB, $\sim18$ msec vs. $\sim4$ mins for CIFAR10, and $\sim10$ msec vs. $\sim3$ mins for MNIST. The Neural-Cleanse baseline was not performed on Tiny-ImageNet due to its huge computational burden.

\subsection{Generalizability of ULPs}

So far, we have shown that the ULPs are capable of detecting poisoned models on unseen poisoning attacks (i.e., unknown triggers), however, for fixed architectures (i.e., known model type). A natural question is how generalizable are ULPs concerning different model architectures?  To that end, we carried out additional experiments on GTSRB. On GTSRB, we trained 300 poisoned and 300 clean models with random VGG-like architectures and where we enforced randomness of networks by randomizing the depth, the number of convolutional kernels, and the number of fully connected units. Also, we trained 200 poisoned and 200 clean models with random ResNet18-like architectures where we, similarly, enforced randomness in the depth and the number of convolutional kernels.

\begin{figure}
    \centering
    \includegraphics[width=.97\columnwidth]{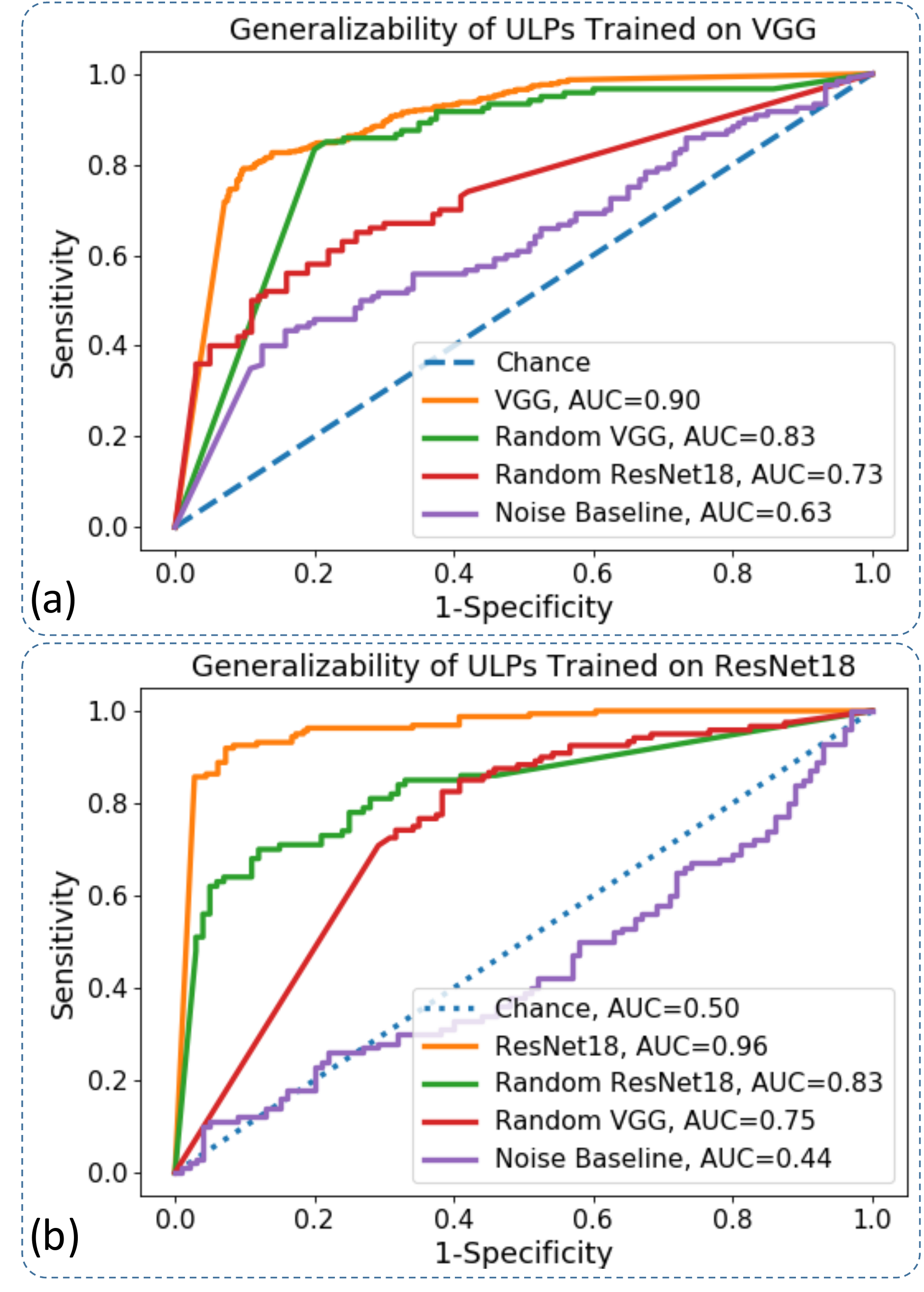}
    \caption{Generalizability of ULPs to random architectures on the GTSRB dataset. The generalizability of the noise baseline and the $M=10$ ULPs trained on a fixed VGG architecture (a) and a fixed ResNet architecture (b) to random VGG and ResNet architectures.}
    \label{fig:generalizability}
     \vspace{-.1in}
\end{figure}

We then tested the trained ULPs (trained on a fixed architecture) on the randomized architectures. Figure \ref{fig:generalizability} shows the generalizability results of the ULPs on these random models, where we also include the ROC curves of the fixed architecture for ease of comparison. The ULPs trained on fixed models remain to be generalizable to random architectures. Moreover, interestingly, we observed that the ULPs trained on a fixed VGG or ResNet architecture are also generalizable to other architecture types, albeit with some sensitivity loss; so, training ULPs on the same architecture type as used for detection is preferable.   %
Finally, to measure the generalizability of the networks across different sizes of triggers on the GTSRB dataset, we trained ULPs (with M=5) on poisoned models with $7\times 7$ triggers. Then, we tested the ULPs on detecting poisoned models containing $5\times 5$ trigger attacks. The ULPs successfully identified the poisoned models with 0.83 AUC. While generalizing from $5\times 5$ to $7\times 7$ resulted in $0.80$ AUC.

\subsection{Adaptive Attacker}
\begin{figure}
    \centering
    \includegraphics[width=.93\columnwidth]{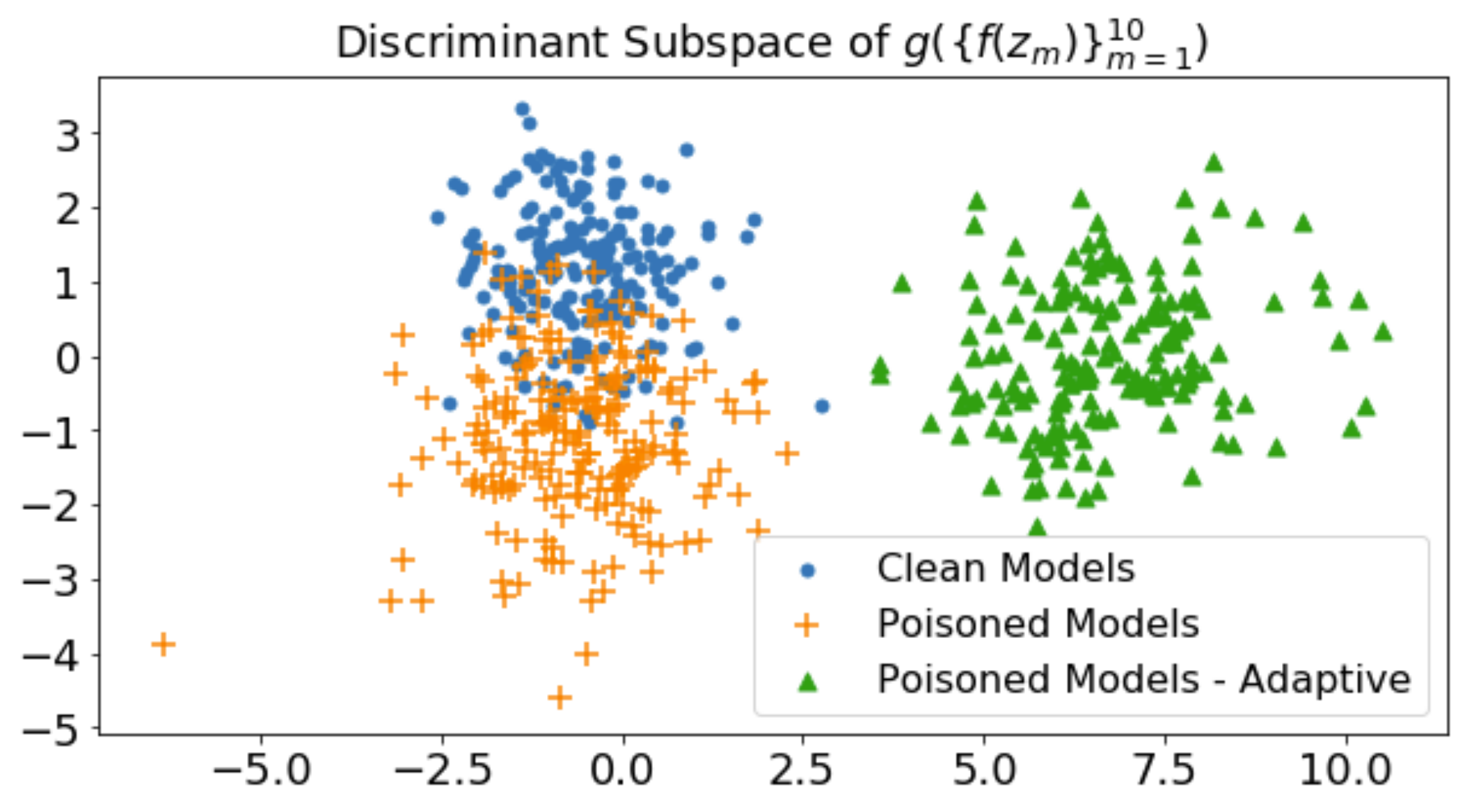}
    \caption{Distribution of the pooled output of clean, poisoned, and adaptively poisoned models for $M=10$ input ULPs in the discriminant subspace. }
    \label{fig:adaptive}
    \vspace{-.2in}
\end{figure}

We also conducted experiments with adaptive adversaries where the attacker has full access to the ULPs and the corresponding binary classifier.  The attacker then regularizes the poisoning loss with the cross-entropy of the detector output and the one-hot vector of the clean class to fool the ULP detector. Unsurprisingly, we observed that the adaptively trained poisoned models could successfully bypass the ULP detector, though this type of full-access attack is often impractical. Interestingly, however, we found that the response of the models remained to be highly discriminant of clean, poisoned, and adaptively poisoned classes: Figure \ref{fig:adaptive}  shows the distribution of the pooled response for these models. This experiment suggests that the ULP defense can be hardened against adaptive attacks, for instance, by increasing the complexity of the binary classifier or utilizing more advanced model augmentations.

\section{Discussion}

We introduced a new method for detecting backdoor attacks in neural networks: Universal Litmus Patterns. The widespread use of downloadable trained neural network increases the risk of working with malicious poisoned networks: networks that were trained such that a visual trigger within an image causes a targeted or untargeted misclassification. So, there is a need for an efficient means to test if a trained network is clean.

Our ULPs are input images that were optimized on a given set of trained poisoned and clean network models, $\{f_n\}$. Here, we need access only to the input-output relationship of these models. So, our approach is agnostic to the network architecture. Moreover, in contrast to prior work, we do not need access to the training data itself.

Surprisingly, our results show that a small set ($\le 10$) of ULPs was sufficient to detect malicious networks with relatively high accuracy, outperforming our baseline, which was based on Neural Cleanse \cite{wang2019neural}. Neural Cleanse is computationally expensive since it requires testing for all possible input-output class-label pairs. In contrast, each ULP requires only one forward pass through a CNN.

We tested ULPs on a trigger set that was disjoint from the set used for optimization and on models different from the source models. We showed generalizability to new triggers as well as new architectures (i.e., random architectures).

Our intuition for why ULPs work for detection is as follows: CNNs essentially learn patterns that are combinations of salient features of objects, and a CNN is nearly invariant to the location of these features. When a network was poisoned, it learned that a trigger is a key feature of a certain object. During our optimization process, each ULP is formed to become a collection of a wide variety of triggers. So, when presenting such a ULP, the network will respond positively with high probability if it was trained with a trigger. In future work, we will investigate ways to harden ULP-based detection against adaptive attacks.\\

\noindent {\bf Acknowledgement:} This work was funded in part under the following financial assistance awards: 60NANB18D279 from U.S. Department of Commerce, National Institute of Standards and Technology, funding from SAP SE, and also NSF grant 1845216.

\newpage
{\small
\bibliographystyle{ieee_fullname}
\bibliography{ULPs}
}

\end{document}